\documentclass[10pt,journal,compsoc]{IEEEtran}

\ifCLASSOPTIONcompsoc
\usepackage[nocompress]{cite}
\else
\usepackage{cite}
\fi

\usepackage{amsmath,amssymb,amsfonts}
\usepackage{graphicx}
\usepackage{textcomp}
\usepackage{xcolor}
\def\BibTeX{{\rm B\kern-.05em{\sc i\kern-.025em b}\kern-.08em
		T\kern-.1667em\lower.7ex\hbox{E}\kern-.125emX}}

\usepackage[ruled, vlined, linesnumbered]{algorithm2e}

\usepackage[justification=centering,listofformat=subsimple,labelformat=simple]{subfig}

\usepackage{caption}
\usepackage{makecell}

\usepackage{etoolbox}
\makeatletter
\patchcmd{\@makecaption}
{\scshape}
{}
{}
{}
\makeatother

\usepackage{multirow}
\usepackage{tabularx}
\usepackage{array}
\usepackage{soul}
\usepackage{hyperref}

\renewcommand{\[}[0]{\textnormal{[}}
\renewcommand{\]}[0]{\textnormal{]}}

\newcommand{\PreserveBackslash}[1]{\let\temp=\\#1\let\\=\temp}

\newcolumntype{C}[1]{>{\PreserveBackslash\centering}p{#1}}
\newcolumntype{R}[1]{>{\PreserveBackslash\raggedleft}p{#1}}
\newcolumntype{L}[1]{>{\PreserveBackslash\raggedright}p{#1}}

\newcommand{\pluseq}{\mathrel{+}=}
\newtheorem{theorem}{Theorem}
\newtheorem{property}{Property}

\begin{document}
	
	\title{Accelerating Backward Aggregation in GCN Training with Execution Path Preparing on GPUs}
	
	\author{Shaoxian~Xu,
		Zhiyuan~Shao,~\IEEEmembership{Member,~IEEE,}
		Ci~Yang,
		Xiaofei~Liao,
		and~Hai~Jin,~\IEEEmembership{Fellow,~IEEE}% <-this % stops an unwanted space
		\IEEEcompsocitemizethanks{
			%			\IEEEcompsocthanksitem The authors are with the National Engineering Research Center for Big Data Technology and System, Services Computing Technology and System Lab, Cluster and Grid Computing Lab, School of Computer Science and Technology, Huazhong University of Science and Technology, Wuhan, 430074, China.\protect\\E-mail: \{sxxu, zyshao, yangci, xfliao, hjin\}@hust.edu.cn
			\IEEEcompsocthanksitem Shaoxian Xu and Zhiyuan Shao are with the National Engineering Research Center for Big Data Technology and System/Services Computing Technology and System Lab/Cluster and Grid Computing Lab, School of Computer Science and Technology, Huazhong University of Science and Technology, Wuhan, 430074, China, and also with Zhejiang Lab, Hangzhou, 311121, China. E-mail: \{sxxu, zyshao\}@hust.edu.cn.
			\IEEEcompsocthanksitem Ci Yang, Xiaofei Liao, and Hai Jin are with the National Engineering Research Center for Big Data Technology and System/Services Computing Technology and System Lab/Cluster and Grid Computing Lab, School of Computer Science and Technology, Huazhong University of Science and Technology, Wuhan, 430074, China. E-mail: \{yangci, xfliao, hjin\}@hust.edu.cn.
			%\IEEEcompsocthanksitem J. Doe and J. Doe are with Anonymous University.
		}% <-this % stops an unwanted space
		\thanks{This work was supported by the National Natural Science Foundation of China under Grant No. 61972444, 61825202, 62072195, and 61832006. This work was also supported by Zhejiang Lab (Grant No. 2022P10AC02).\protect\\(Corresponding author: Zhiyuan Shao.)}
	}
	
	\markboth{Journal of \LaTeX\ Class Files,~Vol.~14, No.~8, August~2015}%
	{Shell \MakeLowercase{\textit{et al.}}: Bare Demo of IEEEtran.cls for Computer Society Journals}
	
	\IEEEtitleabstractindextext{
		\begin{abstract} 
			The emerging \textit{Graph Convolutional Network} (GCN) has been widely used in many domains, where it is important to improve the efficiencies of applications by accelerating GCN trainings. Due to the sparsity nature and exploding scales of input real-world graphs, state-of-the-art GCN training systems (e.g., GNNAdvisor) employ graph processing techniques to accelerate the message exchanging (i.e. aggregations) among the graph vertices. Nevertheless, these systems treat both the aggregation stages of forward and backward propagation phases as all-active graph processing procedures that indiscriminately conduct computations on all vertices of an input graph. 
			In this paper, we first point out that in a GCN training problem with a given training set on an input graph, its aggregation stages of backward propagation phases (called as \textit{backward aggregations} in this paper) can be equivalently converted to partially-active graph processing procedures, which conduct computations on only partial vertices of the input graph. By leveraging such a finding, we propose an execution path preparing method that collects and coalesces the graph data used during different training layers of backward aggregations, and constructs their corresponding sub-graphs (called as \textit{execution paths} in this paper) as inputs to conduct the backward training on GPUs. Further, we propose a structural-aware strategy for the execution paths to compute their optimal group sizes, so as to gain as high as possible performances on GPUs during the backward aggregations. The experiment results by conducting GCN training in typical real-world graphs show that compared with GNNAdvisor, our approach improves the performance of backward aggregations by up to 5.68x on NVIDIA P100 GPU, and up to 6.57x on NVIDIA V100S GPU.
		\end{abstract}
		
		}
	
	\maketitle
	
	\IEEEdisplaynontitleabstractindextext
	
	\IEEEpeerreviewmaketitle
	
	\setlength{\floatsep}{3pt}
	\setlength{\textfloatsep}{3pt}
	\setlength{\skip\footins}{7pt}
	\IEEEraisesectionheading{\section{Introduction}\label{sec:introduction}}
	
	\IEEEPARstart{B}{y} taking a real-world graph as input and conducting convolution operations in it, \textit{Graph Convolutional Network} (GCN)~\cite{kipf:gcn} is extensively used in many fields to extract interesting findings. For example, GCN in recommendation systems~\cite{rex:Recommender} recommends products of interest to the users based on their browsing and purchasing history; GCN in traffic predictions~\cite{zheng:Traffic} helps drivers make the most time-efficient route; GCN can also be used in stock price prediction~\cite{chen:stock} helping people get a head start in the stock market; GCN in the field of medicine helps people to discover new antibiotics~\cite{stokes:antibiotic}, predict protein functions~\cite{gligorijevic:protein}, predict drug-drug interactions~\cite{lin:KGNN}, etc.
	
\begin{figure}[!t]  
		\centering
		\captionsetup[subfloat]{captionskip=2pt}
		\setlength{\abovecaptionskip}{3pt}
		\subfloat[Overall GCN training breakdown]{\includegraphics[width=0.49\linewidth]{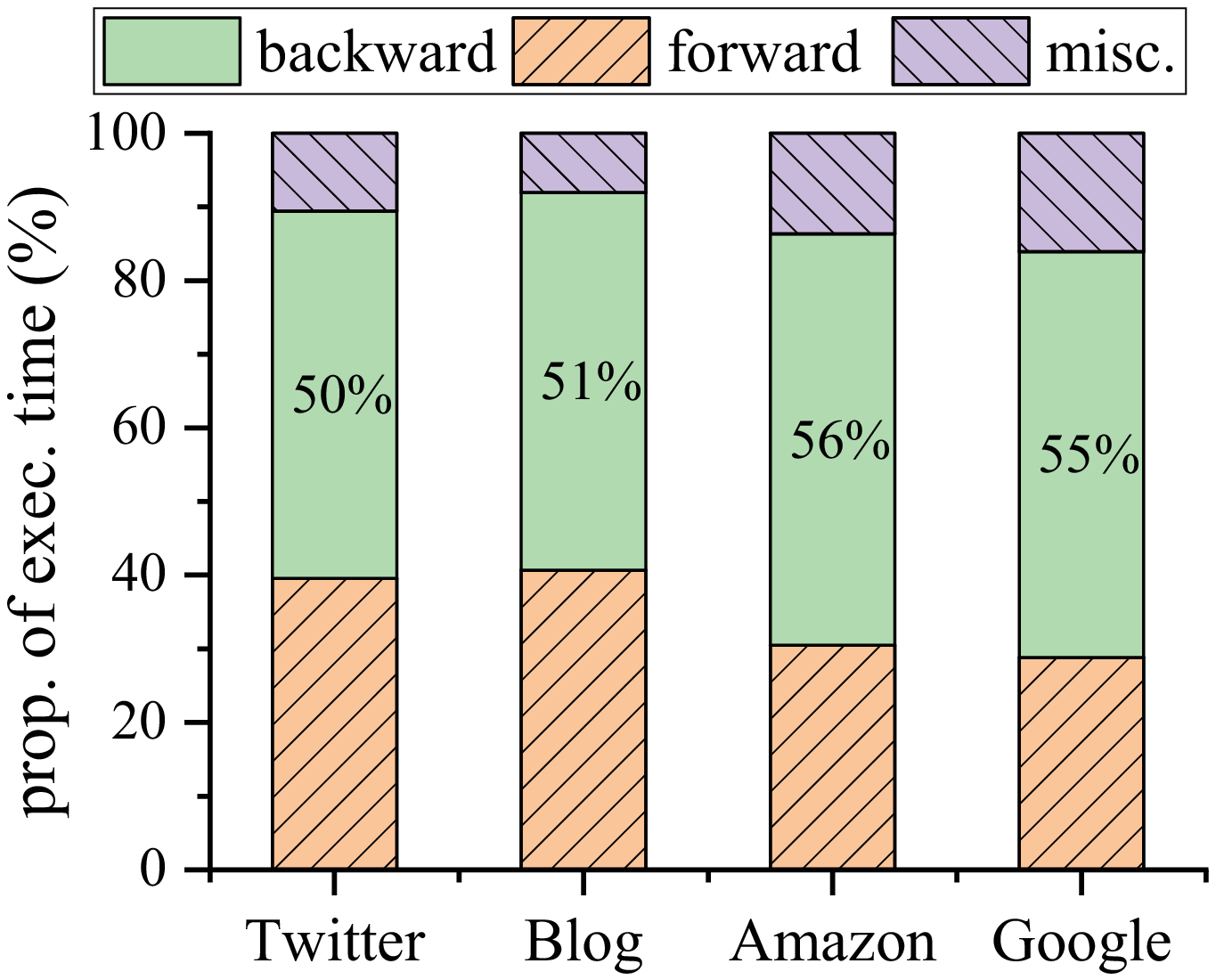}
		}
		\subfloat[Backward propagation breakdown]{\includegraphics[width=0.49\linewidth]{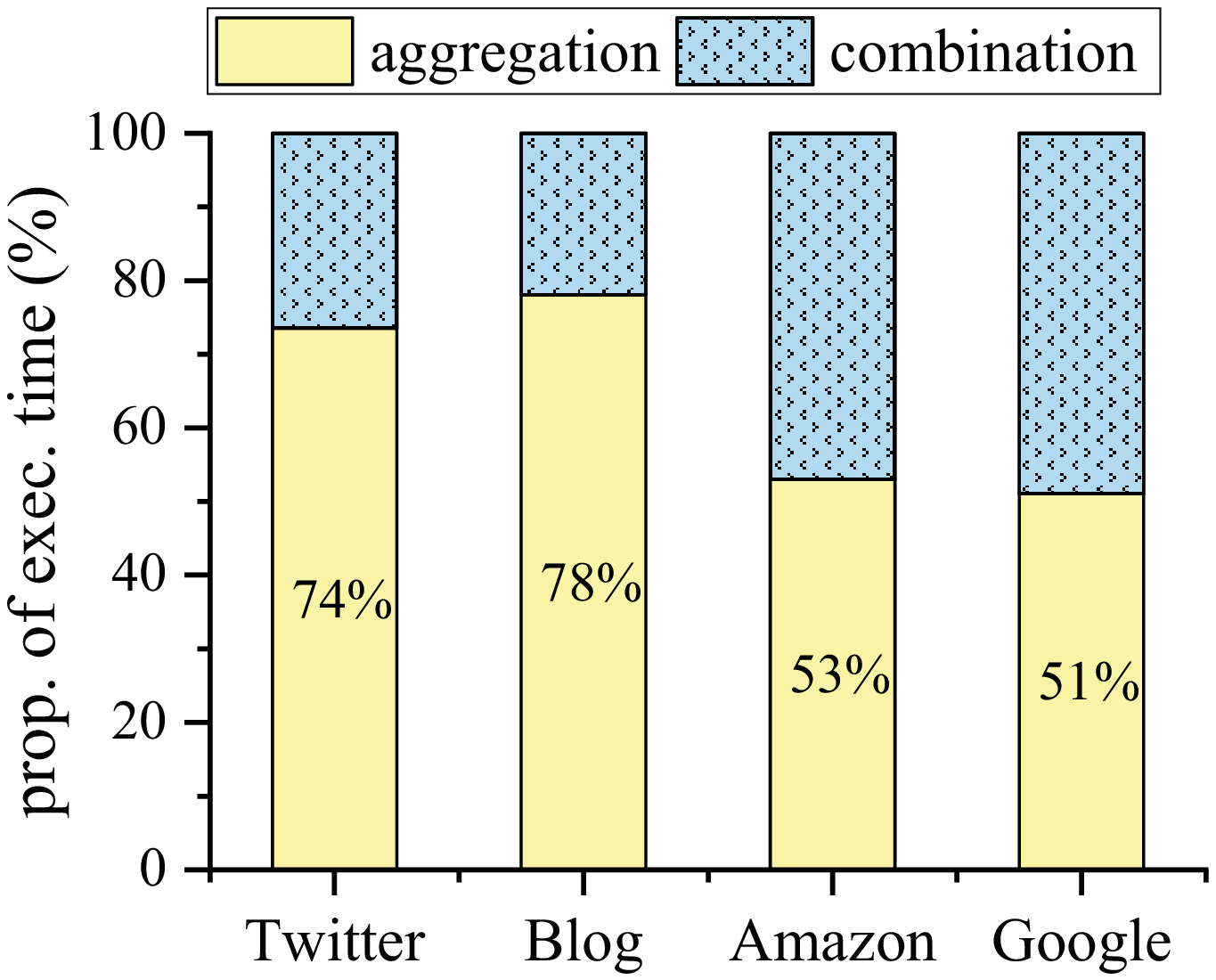}
		}
		\caption{GCN training time breakdown\protect\footnotemark }
		\label{fig:timebreak}
	\end{figure}

	Before using GCN in an application, a corresponding GCN model needs to be trained. The GCN training is time-consuming as an iterative process that typically consists of hundreds or thousands of \textit{epochs}, each of which mainly contains two phases: a \textit{forward propagation} and a \textit{backward propagation}. To study the overheads of different processing phases in GCN training, Figure \ref{fig:timebreak} breaks down the training times on four typical real-world graphs taken from Table 3. From Figure \ref{fig:timebreak}, we can observe that the backward propagation phase typically occupies about half (50\%$\sim$56\%) of the GCN training time, and in the backward propagation time, the aggregation stage (message exchanging among vertices) constitutes the major part of 51\%$\sim$78\%. Hence, it is lucrative to improve the efficiency of the \textit{backward aggregation}\footnotemark stage to accelerate the overall GCN training conducted in real-world graphs, and the theoretical upper bound on overall performance improvement can be up to 1.66x (assuming the backward aggregation times $\approx$ 0). 
	
	Considering the sparsity nature and exploding scales of input real-world graphs, today's state-of-the-art GCN training systems employ graph processing techniques to accelerate both the forward and backward aggregation stages. For example, compared with PyTorch~\cite{adam:PyTorch} which conducts aggregations based on matrix operations~\cite{liu:g3}, GNNAdvisor~\cite{wang:GNNAdvisor} improves the efficiency of aggregation by about 5x on GPUs. However, all these existing systems treat both the forward aggregation and backward aggregation as all-active graph processing procedures that conduct computation indiscriminately on all vertices of an input graph.
	
	\footnotetext{For brevity of discussions in this paper, we call the aggregation stage of backward propagation phase as backward aggregation, and equivalently, that of forward propagation as forward aggregation.} 
	
	In this paper, we show that although the forward aggregation is all-active by the nature of GCN training, \textit{the backward aggregation can be equivalently converted to a partially-active graph processing procedure}, which needs only partial graph data during computation. We leverage such a finding to further optimize the backward aggregation on GPUs (the mainstream vehicles for GCN training), by proposing an \textit{execution path preparing} approach. Before the invocations of backward propagation, our approach first extracts the graph data (vertices and edges of the input graph) to be accessed by each layer of the GCN model, and then constructs corresponding sub-graphs (called as \textit{execution paths} in this paper), in which the backward aggregations are conducted (instead of the original input graph).  
	
	As the portions of graph data accessed during backward aggregation differ from one training layer to another, the execution paths (i.e., extracted sub-graphs in each training layer) produced by our approach differ from each other. The backward aggregations that are conducted in \textit{one} input graph of the original GCN training problem are now conducted in \textit{multiple} sub-graphs. The important parameter (i.e., \textit{group size}) computed from the input graph and leading to optimal performances for GCN trainings conducted on GPUs, does not also lead to optimal performances for the sub-graphs during backward aggregations. We further study the effectiveness of the group sizes on performances of backward aggregation in the sub-graphs, and propose a structure-aware method to compute the group sizes to achieve as high as possible performances for the execution paths. 
	
	\footnotetext{Trainings in this figure are conducted with GNNAdvisor~\cite{wang:GNNAdvisor} on an NVIDIA Tesla P100 GPU~\cite{P100}, while the input graphs are taken from Table \ref{tab:dataset}. In each training, each vertex is given a 128-length random feature vector, and the training set contains 10\% vertices of each input graph.}
	
	Our proposed approach not only reduces the amount of computation during backward aggregation, but also overcomes the performance problems (e.g., branch divergence ~\cite{timothy:Divergence-aware}) on GPUs caused by straightforward selective scheduling methods (e.g., conditional statements), and leads to 1.48x$\sim$5.68x performance improvements on backward aggregation compared with the state-of-the-art GNNAdvisor\cite{wang:GNNAdvisor}. Moreover, the overall GCN training can benefit from our approach: when using our approach during pre-processing, the performance of overall GCN training is improved by about 1.05x$\sim$1.37x; when using our approach on-the-fly (i.e., along with training), the overall training performance is improved by about 1.03x$\sim$1.35x.
	
	This paper makes the following contributions:
	
	$\bullet$ points out that from the angle of graph processing, the backward aggregation of GCN training can be equivalently converted to a partially-active graph processing procedure.
	
	$\bullet$ proposes an execution path preparing approach to reduce the amount of computation in the training by leveraging the partially-active nature of backward aggregation. 
	
	$\bullet$ studies the effectiveness of group sizes on performances of backward aggregation in different execution paths, and proposes a structure-aware method on computing the optimal group sizes for the execution paths. 
	
	$\bullet$ implements and extensively evaluates our proposed approach by revising GNNAdvisor to study the effectiveness of our approach in GCN trainings. 
	Our code is publicly available at \textit{\url{https://github.com/Catriminal/EPPGCN}}.
	
	The rest of our paper is organized as follow: Section \ref{sec:background} introduces background knowledge about GCN training and graph processing, and discusses the related works of this paper. Then Section \ref{sec:partially_active} proves our finding of the partially-active nature of backward aggregation, and proposes an improved algorithm of backward aggregation. We elaborate our execution path preparing method in Section \ref{sec:path_preparing}, and our structure-aware grouping in Section \ref{sec:structure_aware_grouping}. Section \ref{sec:eval} evaluates the proposed methods, and Section \ref{sec:conclusion} concludes the paper.
	
	\section{Background and Related Works} \label{sec:background}
	
	\subsection{GCN training} \label{subsec:background_GCN}
	
	We list the notations used in explaining GCN training in Table \ref{tab:symbol}, and the training process in Algorithm \ref{algo:training}. The training of a GCN model takes as input an undirected graph $G=<V,E>$ ($V$ stands for the vertex set and $E$ denotes the edge set), a vertex feature matrix $X^{(0)}$, a training set $V_t$ and a reference matrix $R$. The input feature matrix $X^{(0)}$, sized $|V|\times f$ ($f$ is the number of features of a vertex), consists of the feature vectors of all vertices in $V$. The training set $V_t$ ($V_t \subset V$) is the subset of vertices (called \textit{training vertices}) with known labels. And the ratio of training vertices to total vertices (i.e., $|V_t|/|V|$) is known as the \textit{training ratio}. The reference matrix $R$, sized $|V|\times c$ ($c$ is the number of classes of vertices), contains the one-hot vectors for training vertices and zero vectors for others. 
	
	The objective of GCN training is to correct a series of weight matrices $W^{(l)}$s to improve the inference \textit{Accuracy}\footnote{``Accuracy'' is a metric for evaluating classification models, and is defined as the number of correct predictions divided by the total number of predictions made by the model.}. A typical GCN model mainly consists of multiple ($L$ in Table \ref{tab:symbol}) convolutional \textit{layers}, at each of which a weight matrix $W^{(l)}$ ($l \in [0, L-1]$) is given. The weight matrices are used to transform input features to obtain classification results. However, since we do not know the relationship between vertex features and vertex categories at the beginning, we have to initialize the weight matrices randomly and conduct a training process. 
	
	The training process is organized into multiple \textit{epochs} (as listed in Algorithm \ref{algo:training}) to correct the weight matrices iteratively, where $max\_epoch$ is generally up to hundreds or thousands. Each of the epochs mainly contains a forward propagation phase and a backward propagation phase. As the names suggest, the forward propagation phase conducts the computation from lower layers to higher layers, while the backward phase follows the reverse direction. 
	
	\begin{table}[!t]
		\centering
		\setlength{\abovecaptionskip}{0pt}
		\setlength{\belowcaptionskip}{5pt}
		\caption{Notations for GCN training}
		\label{tab:symbol}
		\begin{tabular}{m{0.196\linewidth}|m{0.705\linewidth}}
			\hline
			\textbf{Symbol}		&\textbf{Description} \\ \hline
			$G=<V,E>$		&The input undirected graph $G$. $V$ denotes the vertex set, and $E$ denotes the edge set.		\\ \hline
			$e_{uv}$    	&The weight of the edge $u \rightarrow v$.	\\ \hline
			$V_t$			&The input training set (a subset of $V$).			\\ \hline
			$f$				&The number of input features of each vertex.						\\ \hline 
			$X^{(0)}$		&The $|V|\times f$ feature matrix of the input graph, whose $v^{th}$ row is the feature vector of vertex $v$.									\\ \hline
			$c$				&The number of classes of vertices.								\\ \hline
			$R$		&The input reference $|V|\times c$ matrix, whose $v^{th}$ row is a one-hot vector to mark which class vertex $v$ belongs to. \\ \hline 
			$L$				&The number of layers of the GCN model, and we take $L=2$ in this paper.	\\ \hline		
			$dim_l$         &The number of vertex features generated in forward propagation of Layer $l$. We take $dim_0=16$ and $dim_1=c$ in this paper, following the GCN paper~\cite{kipf:gcn} and GNNAdvisor~\cite{wang:GNNAdvisor}.\\ \hline
			$W^{(l)}$		&The (randomly-initialized) weight matrix of Layer $l$. $W^{(0)} \in\mathbb{R}^{f\times dim_0}$ and $W^{(1)} \in\mathbb{R}^{dim_0\times dim_1}$. \\ \hline		
			$Y^{(l)}$		& The result matrix of the forward aggregation of Layer $l$.  \\ \hline 
			$Z$			    &The output result $|V|\times c$ matrix, where the value of the element at $v^{th}$ row and $j^{th}$ column represents the probability that vertex $v$ belongs to class $j$.				\\ \hline 
			$W^{(l)T}$		& Transpose matrix, the same for $Y^{(l)T}$.	\\ \hline
			$W^{(l)'}$		& Derivative matrix, also known as the gradient matrix, the same for $Y^{(l)'}$ and $X^{(l)'}$.	\\ \hline
			$\mathcal{N}_{V_t}^{k}$  & The set of $k$-neighbors of vertices in $V_t$, $\mathcal{N}_{V_t}^{0}=V_t$. \\ \hline
			$\partial$		& Partial derivative. \\ \hline
			$\odot$			& Scalar-Matrix multiplication. \\ \hline
			$\otimes$		& Matrix-Matrix multiplication. \\ \hline
		\end{tabular}
	\end{table}

	During the forward propagation at Layer $l$, each vertex $v\in V$ accumulates (by vector addition) the information from all its neighbors (line 2$\sim$5 of Algorithm \ref{algo:training}), where the information of a neighbor vertex (say $u$) is computed by multiplying the weight of the edge $e_{uv}$ with $u$'s feature vector $X_u^{(l)}$. Such an accumulation stage is known as the ``Aggregation'' in GCNs. Then a ``Combination'' stage (line 6) is conducted to multiply the aggregation result matrix $Y^{(l)}$ with the weight matrix $W^{(l)}$, and apply an activation function $\sigma(\cdot)$\footnote{We use the \textit{ReLU} function and \textit{Softmax} function~\cite{kipf:gcn,wang:GNNAdvisor} as the activation functions for Layer 0 and Layer 1 respectively.} on the multiplication result to obtain the output feature matrix $X^{(l+1)}$ of Layer $l$. After the forward propagation phase, $Loss\_computing(\cdot)$ function computes the $loss$, i.e., the distance value between the known classification results (stored in $R$) of vertices in $V_t$ and their predicted results (stored in $X^{(L)}$). $Grad\_computing(\cdot)$ function then computes the partial derivative of $loss$ with respect to $X^{(L)}$, and the derivative result (known as the \textit{gradient}) is stored in the matrix $X^{(L)'}$, which will be used in the backward propagation phase. 
	
	\begin{algorithm}[t]
		\caption{GCN training (all-active version)}
		\label{algo:training}
		\LinesNumbered
		\SetKw{IN}{in}
		
		\KwIn{$G(A)$, $V_t$, $f$, $X^{(0)}$, $c$, $R$, $L$, $W^{(l)}$, $max\_epoch$}
		\KwOut{$Z$, $W^{(l)}$ of each layer}
		
		\For{$epoch$ \IN \[0, 1, ..., $max\_epoch$-1\]}{
			\tcp{\textbf{Forward propagation}}
			\For{$l$ \IN \[0, 1, ..., $L$-1\]}{
				\ForEach(\tcp*[h]{\textbf{Aggregation}}){$v$ \IN $V$}{		
					\ForEach{$u$ \IN $N(v)$}{
						$Y_v^{(l)} \pluseq e_{uv}\odot X_u^{(l)}$\;
					}
				}
				$X^{(l+1)} = \sigma(Y^{(l)}\otimes W^{(l)})$;\tcp{\textbf{Combination}}
			}
			$loss = Loss\_computing(X^{(L)}, R, V_t)$\;
			$X^{(L)'} = Grad\_computing(X^{(L)}, R, loss)$\;
			\tcp{\textbf{Backward propagation}}
			\For{$l$ \IN \[$L$-1, $L$-2, ..., 0\]}{
				$W^{(l)'} = Y^{(l)T}\otimes X^{(l+1)'}$;\tcp{\textbf{Combination}}
				$Y^{(l)'} = X^{(l+1)'}\otimes W^{(l)T}$;\tcp{\textbf{Combination}}
				\ForEach(\tcp*[h]{\textbf{Aggregation}}){$v$ \IN $V$}{		
					\ForEach{$u$ \IN $N(v)$}{
						$X_v^{(l)'} \pluseq e_{vu}\odot Y_u^{(l)'}$\;
					}
				}
			}
			$Param\_updating$($epoch$, $W^{(0)}$, $W^{(0)'}$ ... $W^{(L-1)}$, $W^{(L-1)'}$)\;
		}
		$Z = X^{(L)}$\;
	\end{algorithm}
	
	Subsequently, a backward propagation phase is conducted to compute the series of gradient matrices (i.e., $W^{(l)'}$s) for all training layers. At layer $l$, it first combines the derivative values in $X^{(l+1)'}$ and the aggregation result (i.e., $Y^{(l)T}$, which is the transpose of $Y^{(l)}$) from the forward propagation to compute the gradient of the weight matrix $W^{(l)}$. To further compute the weight matrix of the next (i.e., $l-1$, recall that we are going in a backward direction) layer, the backward propagation first computes an intermediate gradient matrix of $Y^{(l)'}$, and then uses it to compute the derivative matrix of $X^{(l)'}$. The method on computing $X^{(l)'}$ is to invoke another aggregation stage (i.e., the for-loop in line 12$\sim$14, called as \textit{backward aggregation} in this paper), during which each vertex of the input graph accumulates information from all its neighboring vertices. Finally, $Param\_updating(\cdot)$ function employs the Adam algorithm \cite{diederik:adam} to implement the gradient descent~\cite{ruder:gradient_descent} on the weight matrices of $W^{l}$s by using the gradient matrices $W^{(l)'}$s computed during backward propagation, such that later epochs of the training will have smaller losses.

	\subsection{Graph processing systems on GPUs} \label{subsec:background_GP}
	
	Graph processing on GPUs considers basic graph algorithms (like PageRank~\cite{shi:PageRank} and Breadth First Search~\cite{liu:bfs}) in graphs that are much larger than the size of L1 cache, but smaller than that of the GPU global memory. Popular graph processing systems on GPUs adopt the vertex-centric programming model~\cite{shi:GPUsurvey} that conducts a graph algorithm by iteratively executing its corresponding function over the vertices of an input graph. The algorithm's corresponding function typically declares how data are transmitted along the incident edges of a vertex. Regarding the direction of data transmission, there are two modes: \textit{push} and \textit{pull}. 
	
	In push mode, a vertex pushes its messages to its neighbors and updates the properties of the neighbors. In pull mode, a vertex pulls the properties of its neighbors and updates itself. The push mode leads to huge atomic overheads during a highly paralleled execution~\cite{Besta:push_pull}, since different vertices (threads) may push (write) their data to the same vertex simultaneously. Whereas, the pull mode needs to traverse extra edges than necessary if only part of the vertices hold useful data~\cite{Besta:push_pull}. 
	
\begin{figure}[t]
		\centering
		\setlength{\abovecaptionskip}{3pt}
		\includegraphics[width=0.99\linewidth]{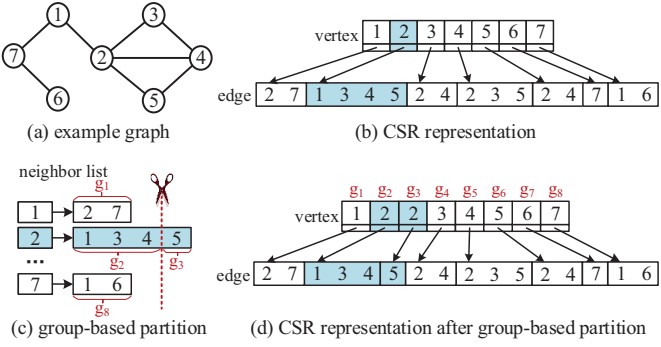}
		\caption{Example of neighbor grouping, assuming \textit{group size}=3.}
		\label{fig:group_part}
		%	\vspace{-15pt}
	\end{figure}

	To balance the load among thousands of hardware threads of a GPU, modern graph processing systems (e.g., Tigr~\cite{sabet:Tigr}) divide the neighbor lists into multiple groups (we call this method as \textit{neighbor grouping} in this paper), whose size is subject to an upper-bound value (i.e., \textit{group size}), and distribute the groups among participating software threads. By assuming the group size equals 3, Figure \ref{fig:group_part} gives an example graph in Figure \ref{fig:group_part}(a), and divides the neighbor lists, whose sizes are larger than 3, in its CSR representation (in Figure \ref{fig:group_part}(b)) into multiple groups (in Figure \ref{fig:group_part}(c-d)). In this example, as the length of vertex $v_2$'s neighbor list is bigger than 3, its neighbor list is thus being divided into two groups (i.e., $g_2$ and $g_3$). During processing, different groups will be assigned to different hardware threads of the GPU. Although the neighbor grouping technique also introduces atomic overheads when writing the results of vertices back to the global memory of GPU (e.g., write back the result of $v_2$, while $g_2$ and $g_3$ are distributed two software threads running on two different hardware threads in our example), such atomic overheads can be much lower than that of push mode when the group size is appropriately chosen and sufficiently large.
	
	\textbf{Use of graph processing in GCN training:} 
	State-of-the-art GCN training systems on GPUs~\cite{wang:GNNAdvisor,ma:NeuGraph,huang:gaps} widely adopt techniques (such as vertex-centric programming and neighbor grouping) used in graph processing systems to improve their efficiencies on aggregation. Nevertheless, in GCN training, the number of vertex features can be up to hundreds, which is much larger than that in basic graph algorithms, and thus leads to high atomic overheads if the push mode is used. The pull mode is hence widely employed by the GCN training systems on GPUs. Besides, different from the graph processing systems that generally invoke one hardware thread to conduct the processing of a vertex, GCN training systems typically divide the features of a vertex into multiple fragments, and assign them to different hardware threads for parallel processing. For load-balancing, the neighbor grouping technique used in graph processing is also used by recent GCN training systems on GPUs~\cite{wang:GNNAdvisor,huang:gaps}. Similarly, the use of neighbor grouping also leads to atomic overheads on updating the data attached to the vertices as in graph processing systems. 
	
	\subsection{Related works} \label{subsec:related}
	
	By leveraging the technologies of graph processing, GCN training systems on GPUs~\cite{liu:g3,wang:GNNAdvisor,ma:NeuGraph,huang:gaps} have exhibited significant performance improvements over popular deep learning frameworks such as PyTorch~\cite{adam:PyTorch} and TensorFlow~\cite{mart:Tensorflow}. G$^3$~\cite{liu:g3} proposed to use Gunrock~\cite{wang:gunrock}, a popular graph processing framework on GPU, to conduct GCN training. NeuGraph~\cite{ma:NeuGraph} proposes a new programming model named as Scatter-ApplyEdge-Gather-ApplyVertex by combing the dataflow model and vertex-centric model to support scalable parallel \textit{Graph Neural Network} (GNN) training. The work in \cite{huang:gaps} accelerates the GNN inference by improving the data locality and load balance. GNNAdvisor~\cite{wang:GNNAdvisor} employs a vertex reorder method (i.e., Rabbit~\cite{arai:rabbit}) to improve the data locality, and proposes a 2D workload management method to balance the workloads at vertex level and vertex feature level. Besides, the python libraries for GNNs, such as DGL~\cite{wang:DGL} and PyG~\cite{matthias:PyG} built on top of existing deep learning frameworks like PyTorch, are leveraging state-of-the-art \textit{Sparse Matrix Multiplication} (SpMM) and graph-processing-like techniques (such as representing matrices with vertex arrays and edge lists) to improve their performance.
	
	All these existing GCN training systems still treat both forward and backward aggregations as all-active graph processing procedures, leading to excessive computations during the backward aggregation.

	\section{Partially-active Backward Aggregation}\label{sec:partially_active}
	
	In this section, we focus our discussions on the backward aggregation of GCN training, and show that it can be regarded as an equivalent partially-active graph processing procedure where only a fraction of the graph data is needed. To demonstrate this finding, we first scrutinize the $Loss\_computing(\cdot)$ and $Grad\_computing(\cdot)$ functions, and investigate how the gradients in $X^{(L)'}$ are computed (in line 8 of Algorithm \ref{algo:training}). 
	
	In line 7 of Algorithm \ref{algo:training}, $loss$ is a scalar value computed in Equation \ref{equ:loss}:
	\begin{equation} \label{equ:loss}
	\begin{aligned}
	loss&=Loss\_computing(X^{(L)}, R, V_t)\\
	&=-\sum_{v\in V_t}\frac{1}{|V_t|}\sum_{j=0}^{c-1}R_{v,j}\cdot logX^{(L)}_{v,j}
	\end{aligned}
	\end{equation} 
	
	\noindent
	where $|V_t|$ is the number of the training vertices in $V_t$, $R$ is the reference matrix sized $n\times c$ (where $n$ is the number of vertices of input graph and $c$ is the number of classes of the vertices), and $X^{(L)}$ is the output of forward propagation also sized $n\times c$. From Equation \ref{equ:loss}, we can observe that \textit{only vertices belonging to $V_t$ participate in the computing of $loss$}. 
	
	In gradient matrix $X^{(L)'}$, each element is the partial derivative of $loss$ with respect to its corresponding element in $X^{(L)}$. The $Grad\_computing(\cdot)$ function computes the element at the $v^{th}$ row and $j^{th}$ column of $X^{(L)'}$ by using Equation \ref{equ:derivative} listed in the following.
	
	\begin{equation} \label{equ:derivative}
	\begin{aligned}
	X_{v,j}^{(L)'}	= \frac{\partial loss}{\partial X_{v,j}^{(L)}} 
	= -\frac{\partial\ \sum_{u\in V_t}\frac{1}{|V_t|}\sum_{k=0}^{c-1}R_{u,k}\cdot logX^{(L)}_{u,k}}{\partial\ X^{(L)}_{v,j}}
	\end{aligned}
	\end{equation} 
	
	From Equation \ref{equ:derivative}, we observe that for any vertex $v\in V\setminus V_t$, i.e., vertex belonging to the input graph but \textit{not} in $V_t$, $X_{v,j}^{(L)'}$ will be zero. Since such $X_{v,j}^{(L)}$ does not appear in the computing of $loss$, and consequently, changes in the value of such $X^{(L)}_{v,j}$ do not cause any changes in $loss$. Therefore, we have $\partial loss/\partial X_{v,j}^{(L)}=0$ and the following property for $X^{(L)'}$:
	
	\begin{property}\label{property:XL}
		For $\forall v\in V\setminus V_t$, the elements in the $v^{th}$ row of $X^{(L)'}$ are all zeros.
	\end{property}
	
	Further, we generalize such a property to all the layers in Theorem \ref{thm:property_Xl}, where $\mathcal{N}_{V_t}^{k}$ denotes the set of $k$-neighbors (i.e., if there exists a path from $v$ to $u$, and the path consists of $k$ edges, $u$ is a $k$-neighbor of $v$) of the training vertices in $V_t$:
	
	\begin{theorem}\label{thm:property_Xl}
		For $\forall v\in V\setminus \mathcal{N}_{V_t}^{k}$ ($k\in [0, L]$ and $\mathcal{N}_{V_t}^{0} = V_t$), the elements in the $v^{th}$ row of $X^{(L-k)'}$ are all zeros.
	\end{theorem}
	
	\begin{IEEEproof}
		We prove the theorem by induction, and discuss the case of increasing $k$ from 0 to $L$. For the initial case of $k=0$, Property \ref{property:XL} indicates that for $\forall v\in V\setminus V_t$, the elements in the $v^{th}$ row of $X^{(L)'}$ are all zeros. The theorem thus holds for the initial case. Then we only need to prove that for an arbitrary $k \in [0, L-1]$, the theorem holds for $k+1$, as long as the theorem holds for $k$.
		
		As the theorem holds for $k$, which means that $\forall v\in V\setminus \mathcal{N}_{V_t}^{k}$, the elements in the $v^{th}$ row of $X^{(L-k)'}$ are all zeros. Such a property of $X^{(L-k)'}$ will be transferred to $Y^{(L-(k+1))'}$ by the matrix multiplication (in line 11 of Algorithm \ref{algo:training}). Then in the aggregation stage, for $\forall v$, the row $X_v^{(L-(k+1))'}$ is computed by $X_v^{(L-(k+1))'} \pluseq e_{vu}\odot Y_u^{(L-(k+1))'}$. With the property of $Y^{(L-(k+1))'}$, we know that the row $X_v^{(L-(k+1))'}$ can have nonzero values if and only if $v$ is an immediate neighbor of $\mathcal{N}_{V_t}^{k}$, which means $\forall v\in V\setminus \mathcal{N}_{V_t}^{(k+1)}$, its corresponding row in $X^{(L-(k+1))'}$ must be all zeros.
	\end{IEEEproof}
	
	With Theorem \ref{thm:property_Xl}, we know that \textit{only partial data (the nonzero rows) need to be processed during the backward aggregation of GCN training}, since all-zero rows do not produce useful results in matrix-multiplications when computing the gradient matrix of $W^{(l)'}$s (line 10 of Algorithm 1). 
	
	Based on such a finding, we can leverage the idea of ``selective scheduling'' from the graph processing community to reduce the amount of computing in the aggregation stage of backward propagation, and hence improve the performance of GCN training. In Algorithm \ref{algo:ipv_agg}, we list the partially-active version of backward aggregation. Rather than conducting aggregation operations indiscriminately on all vertices of an input graph as listed in Algorithm \ref{algo:training}, the partially-active version in Algorithm \ref{algo:ipv_agg} confines the scope of aggregation in the $(L-l)$-neighbors of the training set (i.e., $\mathcal{N}_{V_t}^{L-l}$) at Layer $l$, where $l$ decreases from $L-1$ to 0. 
	
	Note that, it does \textit{not} cause any accuracy degradation that replacing the backward propagation phase (line 9$\sim$14) of Algorithm \ref{algo:training} with Algorithm \ref{algo:ipv_agg}. Since Algorithm \ref{algo:ipv_agg} only skips all-zero rows in $Y^{(l)'}$ during the aggregation stage, the backward propagation phase in Algorithm \ref{algo:ipv_agg} can yield exactly the same results (i.e., gradients of the weight matrices) as that in Algorithm \ref{algo:training}. 
	
	\begin{algorithm}[t]
		\caption{partially-active version of backward aggregation in GCN training ($N(v)$ denotes the set of neighboring vertices of $v$, $\mathcal{N}_{V_t}^{k}$ denotes the set of $k$-neighbors for vertices in $V_t$, and $\mathcal{N}_{V_t}^{0}=V_t$)}
		\label{algo:ipv_agg}
		\SetKw{IN}{in}
		
		\KwIn{ $X^{(L)'}$, $Y^{(l)}$s and $W^{(l)}$s produced in forward propagation }
		\KwOut{ $W^{(l)'}$s }
		
		\tcp{\textbf{Backward propagation}}
		\For{$l$ \IN \[$L$-1, $L$-2, ..., 0\]}{
			$W^{(l)'} = Y^{(l)T}\otimes X^{(l+1)'}$;\tcp{\textbf{Combination}}
			$Y^{(l)'} = X^{(l+1)'}\otimes W^{(l)T}$;\tcp{\textbf{Combination}}
			\ForEach(\tcp*[h]{\textbf{Aggregation}}){$v$ \IN $\mathcal{N}_{V_t}^{L-l}$}{
				\ForEach{$u$ \IN $N(v) \cap \mathcal{N}_{V_t}^{L-l-1}$}{
					$X_v^{(l)'} \pluseq e_{vu}\odot Y_u^{(l)'}$\;
				}
			}
		}
	\end{algorithm}
	
	\section{Execution Path Preparing}\label{sec:path_preparing}
	
	Considering the partially-active nature of backward aggregation discussed in Section \ref{sec:partially_active}, it is possible to reduce the overheads of backward aggregation by selectively scheduling. There are two straightforward methods on achieving such an objective: if-else statements and frontier-based approaches. The first method can be implemented with a little effort by simply judging whether a vertex is active (e.g., included in $\mathcal{N}_{V_t}^{l}$ at the $(L-l)^{th}$ training layer during backward aggregation) or not, and conducts computing only on the active vertices. Such a method, however, leads to the branch divergence problem~\cite{timothy:Divergence-aware}, as GPUs are notorious for their inefficiency in handling conditional statements. Moreover, it also leads to uncoalesced memory accesses, as the active vertices and their corresponding neighbor lists are scattered among the vertex array and edge array respectively.

	The frontier-based method is widely employed in graph processing systems, such as Gunrock~\cite{wang:gunrock}, that adopt push mode computation. These systems dynamically build new frontiers (e.g., bitmaps where active vertices are marked as ones) at each iteration, and computations are conducted on active vertices. However, as vertices have rich properties in GCN training problems, pull mode computing is adopted to avoid high performance penalties incurred by atomic operations. Moreover, for a given GCN training problem with a known training set of $V_{t}$, the sets of active vertices at each layer do \textit{not} change during all epochs of the training, which makes it unnecessary to build new frontiers dynamically as in graph processing systems.
	
	Based on the partially-active nature of backward aggregation of GCN training, we propose a simple yet effective approach in this paper, and name it as \textit{execution path preparing}. The basic idea of our proposal is that: for a given graph $G$ and training set $V_t$, we first compute all of the $\mathcal{N}_{V_t}^{k}$s, where $k \in [0, L-1]$. And then, for each layer (say the $l^{th}$ layer), we browse the input graph $G$ to collect all the vertices and their neighbor lists (after being filtered by joining $\mathcal{N}_{V_t}^{L-l-1}$) needed during backward aggregation according to $\mathcal{N}_{V_t}^{L-l}$, and store the collected data in sub-graphs ($SG_{l}$s), also called as "execution paths". In $SG_{l}$, the vertex set is $\mathcal{N}_{V_t}^{L-l}$ and the edge set is $\{N(v) \cap \mathcal{N}_{V_t}^{L-l-1}, \forall v | v \in \mathcal{N}_{V_t}^{L-l}\}$. 
	
	\begin{figure*}[t]
		\setlength{\abovecaptionskip}{3pt}
		\begin{center}
			\begin{minipage}{0.49\textwidth}
				\centering
				\includegraphics[width=0.90\linewidth]{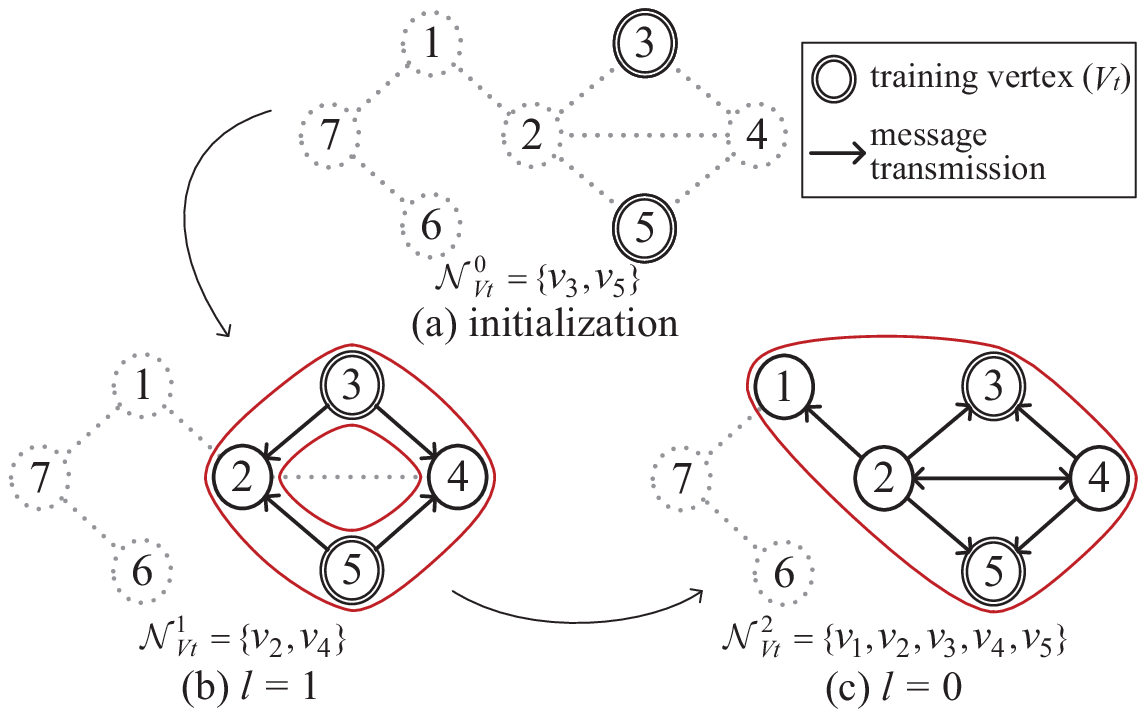}
				\caption{Execution paths of backward aggregations in two layers on the example graph by assuming $V_t=\{v_3, v_5\}$}
				\label{fig:active_edges}
			\end{minipage}
			\hfill
			\begin{minipage}{0.49\textwidth}
				\centering
				\includegraphics[width=0.96\linewidth]{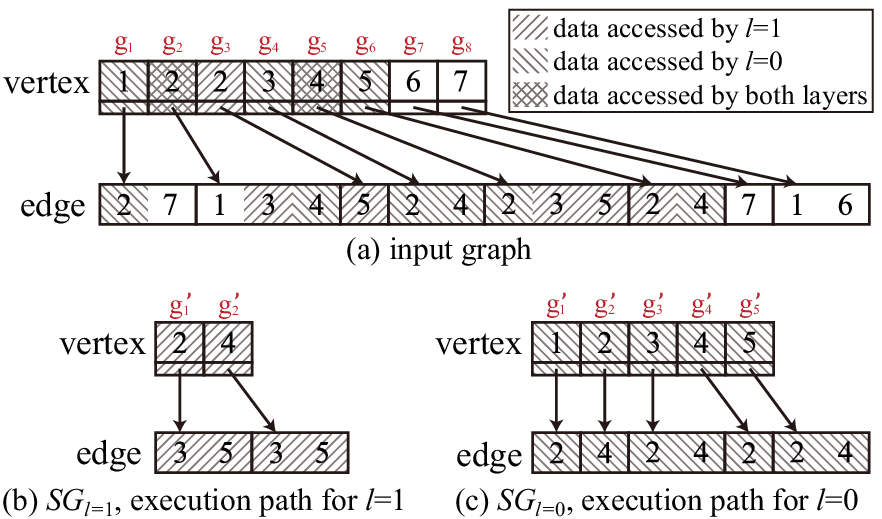}
				\caption{Results of execution path preparing on Figure \ref{fig:active_edges} (vertex array consists of destination vertices as pull mode is adopted)}
				\label{fig:executionpaths}
			\end{minipage}
		\end{center}
		\vspace{-20pt}
	\end{figure*}
	
	We give an example GCN training problem in Figure 3 to explain our proposal. The GCN training problem in Figure \ref{fig:active_edges} takes the training set of $V_t = \{v_3,v_5\}$, and conducts backward aggregation on the example graph illustrated in Figure \ref{fig:group_part}(a). At the first (i.e., $l=1$ as we assume $L=2$) layer of backward aggregation, computations are conducted on vertices in $\mathcal{N}_{V_t}^{1}=\{v_2,v_4\}$, and messages are passed over the edges of $\{v_3\rightarrow v_2, v_5\rightarrow v_2, v_3\rightarrow v_4, v_5\rightarrow v_4\}$. While at the second layer (i.e., $l=0$ as we are going backward) of backward aggregation, computations are conducted on vertices in $\mathcal{N}_{V_t}^{0}=\{v_1,v_2,v_3,v_4,v_5\}$, and messages are passed over the edges of $\{v_2\rightarrow v_1, v_4\rightarrow v_2, v_2\rightarrow v_3, v_4\rightarrow v_3, v_2\rightarrow v_4, v_2\rightarrow v_5, v_4\rightarrow v_5\}$. 
	
	With our execution preparing approach, the execution path of the first layer aggregation (shown in Figure \ref{fig:active_edges}(b)) will be extracted and stored in $SG_1$ whose CSR representation is illustrated in Figure \ref{fig:executionpaths}(b), while the execution path for the second layer aggregation and its corresponding CSR representation is shown in Figure \ref{fig:active_edges}(c) and \ref{fig:executionpaths}(c), respectively. With these execution paths, during the first training layer of backward aggregation, the training will be conducted in sub-graph $SG_1$, rather than in the original input graph $G$, and during the second training layer, $SG_0$ will be used.
	
	Note that, our approach is different from the stochastic training~\cite{william:graphsage,zeng:GraphSAINT}. First, stochastic training uses sub-graphs coming from graph sampling to replace original input graphs to conduct the actual training. The graph sampling generally leads to loss of structural information of the input graph, and hence incurs degradation on the Accuracy~\cite{william:graphsage,jia:acc_loss}. On the contrary, our approach is an optimization technique that can be applied to traditional GCN training, where the sub-graphs constructed by using our approach abide by the actual execution paths of computation of GCN training, and thus do not result in any loss of useful graph structural information and training Accuracy. Second, in stochastic training, sub-graphs are constructed at the beginning of \textit{each} epoch by sampling from the input graph, and used during both forward and backward aggregations. On the contrary, in our approach, sub-graphs are constructed \textit{only once} during pre-processing or the first epoch (on-the-fly) and used only in the backward aggregations.

	\section{Structure-aware Grouping} \label{sec:structure_aware_grouping}
	
	To efficiently use the many-core resource of a GPU, dividing the neighbor list of each vertex into multiple ``groups'' (i.e., neighbor grouping) is a simple yet effective technique used in state-of-the-art graph processing systems (e.g., Tigr~\cite{sabet:Tigr}) as well as GCN training systems (e.g., GNNAdvisor~\cite{wang:GNNAdvisor}). In neighbor grouping, the size of groups is subjected to an upper-bound value, named as group size ($gs$ for short). 
	
	However, when profiling the performance of our execution path preparing, we find that the grouping strategies in state-of-the-art systems cannot find proper $gs$es for our execution paths (i.e., $SG_l$s), and thus lead to severe performance problems (i.e., load unbalance and high atomic overhead). For example, Tigr~\cite{sabet:Tigr} chooses $gs=10$ for all graphs. And GNNAdvisor~\cite{wang:GNNAdvisor} computes the $gs$ for an input graph according to the vertex's feature vector length (i.e., $dim$) as well as the number of hardware threads of the GPU platform. However, in this way, GNNAdvisor assigns the same $gs$ to different graphs on a GPU, as long as the graphs have the same feature vector length.
	
\begin{figure}[t]
		\centering
		\setlength{\abovecaptionskip}{3pt}
		\includegraphics[width=0.8\linewidth]{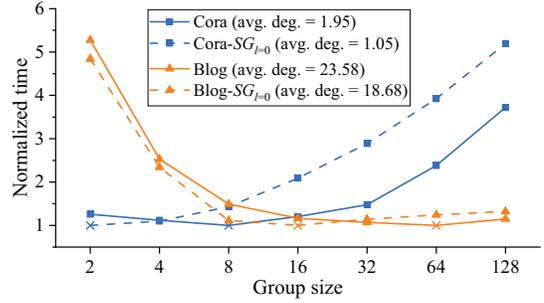}
		\caption{Effectiveness of group size on backward aggregation performance, normalized to the smallest aggregation time (10\% of the vertices are selected in the training set)}
		\label{fig:effect_of_gs}
		%		\vspace{-5pt}
	\end{figure}

	To show that it is not desirable to use the same $gs$ for our sub-graphs (execution paths) as for the original input graph, and to find what factors determine the optimal $gs$ of a (sub-)graph, we choose two real-world graphs from Table \ref{tab:dataset}, and conduct GCN trainings\footnote{Similar to the trainings in Figure \ref{fig:timebreak}, we use the NVIDIA Tesla P100 GPU, and set training ratio $=10\%$, $f=128$, $dim_0=16$ and $dim_1=c$.} by using GNNAdvisor (backward aggregations run in original input graphs) and our approach (backward aggregations run in $SG_l$s) respectively.  Within these experiments, we choose the $gs$ to be 2, 4, 8, ...128, and illustrate the backward aggregation performances of Layer 0 in Figure \ref{fig:effect_of_gs}, where the smallest aggregation times are normalized to ones and marked with $\times$. 
	
	From Figure \ref{fig:effect_of_gs}, we have three observations: 
	
	\noindent
	1) for our chosen graphs, the gap between the best and the worst backward aggregation performances is huge. For example, on \texttt{Blog}, the worst performance appears when $gs=2$, and is 5.3x worse than the best performance (appears when $gs=64$). Such an observation reveals that the choice of $gs$ has a significant impact on the performance of backward aggregation.
	
	\noindent
	2) the patterns of the curves are similar, where there is only one optimal $gs$ for a graph, and smaller or larger $gs$ values lead to worse performances. The reason behind such a phenomenon is that the choice of $gs$ is in-essence a trade-off problem. A small $gs$ benefits load-balancing among the processors (hardware threads of GPUs). However, a smaller $gs$ also leads to a higher chance for a neighbor list of a vertex to be divided into more groups, which increases the atomic overheads on writing back the computing result of each group to the global memory of a GPU. On the contrary, larger group sizes lead to smaller atomic overheads but worse load-balancing.
	
	\noindent
	3) considering the optimal $gs$es among different graphs, we can observe that graphs with higher average degrees generally have larger optimal $gs$ values. The reason behind such a phenomenon is that for graphs with higher average degrees, larger $gs$ values lead to a better trade-off between load-balancing and atomic overheads. Different optimal $gs$es of different graphs also reveal that the structure (e.g., degree distribution) of the graph is a decisive factor to the group size value. Note this observation differs GCN training systems from traditional graph processing systems on GPUs, where a fixed $gs$ works fine as in Tigr~\cite{sabet:Tigr}. Since in GCN training problems, vertices generally have rich properties (e.g., long vectors), the computation on them hence raises much higher atomic overheads than the graph algorithms. 
	
	Considering the third observation from Figure \ref{fig:effect_of_gs} together with our proposed method of execution path preparing which makes the backward aggregations conducted on \textit{multiple} sub-graphs rather than the input graph, a method that can compute optimal $gs$es for different graphs based on their structure information is needed to further improve the efficiency of backward aggregations on GPUs. In this paper, we propose two methods on computing the optimal choice of $gs$ by using not only the $dim$ but also the structure information of the graph.
	
	\textbf{Machine learning based grouping}: 
	In machine learning based grouping, we convert the problem of computing optimal $gs$es for graphs into a graph classification problem~\cite{niepert:graph_class,junhyun:SAG}. The objective of a typical graph classification problem is to identify the classes of graphs (e.g., distinguish whether a graph is a cycle, star, or wheel graph). To achieve such an objective, a graph classification system generally takes as input the graphs as well as the features of vertices in the graphs, and then generates a graph feature vector for each graph according to its vertices' features. Finally, the system infers the classes of the graphs according to their graph features. Besides, similar to GCN training, a subset of input graphs with known classes is used as the training set. 
	
	In order to leverage the graph classification to compute graphs' optimal $gs$es, we take the optimal $gs$es of the graphs as their classes. Thus, the results of graph classification on such graphs are the inferred $gs$es for them. There are two reasons why optimal $gs$es can be used as categories of graphs. One reason is that $gs$ must be a positive integer. The second is that the optimal $gs$ is unique: as discussed in the second observation of Figure \ref{fig:effect_of_gs}, from both the patterns of curves and the trade-off essence of choices on $gs$, it can be seen that the optimal $gs$ for an input graph is unique on a given GPU platform and GCN model.
	
	In this paper, we employ SAGPool-GCN~\cite{junhyun:SAG} (abbreviated as \textit{SAGPG}, one of the state-of-the-art graph classification neural networks) to infer the classes (the optimal $gs$es) of graphs. Based on our third observation on Figure \ref{fig:effect_of_gs}, we take the degree of the vertex as the vertex feature used by SAGPG. Besides, based on GNNAdvisor's method of computing $gs$, we input the $dim$ of the graph to SAGPG as a manual graph feature for the graph. As for the training set, to make the SAGPG with high generalization ability, we use not only the real-world graphs listed in Table \ref{tab:dataset} of Section \ref{sec:eval} but also a set of synthetic graphs generated via PaRMAT~\cite{parmat} as the training set. As for obtaining the reference classification results (i.e., optimal $gs$es) for graphs in the training set, on each graph, we enumerate $gs$ starting from 1 until we find the optimal $gs$. 
	
	After completing the training of SAGPG, when we need to compute the choice of $gs$ for a graph $G$, we just input $G$ as well as its vertices' features (i.e., the degrees of vertices) and graph feature (i.e., $dim$) into the pre-trained SAGPG. And by a single epoch of inference (forward propagation), the pre-trained SAGPG outputs the inferred optimal $gs$ for $G$. 
	In our execution path preparing, different from GNNAdvisor which computes the $gs$ for only one graph, we need to compute multiple (i.e., $L$) $gs$es for different execution paths (i.e., $SG_{l=L-1}$, $SG_{l=L-2}$, ..., $SG_{l=0}$) as we extract a separate execution path for each training layer of the GCN model. We input the multiple execution paths (as well as their corresponding vertex degrees and $dim$s) into the pre-trained SAGPG. And then the SAGPG infers the optimal $gs$ for each inputted execution path.  
	
	\textbf{Linear regression based grouping}:
	In linear regression based grouping, to make full use of the graph structure information for the regression of $gs$, we take into account not only the average degree and the $dim$ of a graph, but also the number of vertices and the number of edges. 
	
	Before building the regression equation, we need to identify the linear relationship between the structure information (as well as the $dim$) and the optimal $gs$ of the graph at first. We conduct the bivariate correlation analysis~\cite{swank:b-correlation} and the partial correlation analysis~\cite{baba:p-correlation} on the metrics we choose (i.e., \# vertices, \# edges, average degree, $dim$, and optimal $gs$). The analysis results are reported in Table \ref{tab:correlation}, where $r$ is the Pearson correlation coefficient between two metrics, and $p$-value can be regarded as the probability that the correlation shown in the data used in the analysis occurs by chance. The closer the absolute value of $r$ is to 1, the stronger the linear correlation between the two metrics. And the correlation is considered statistically significant (i.e., considered to be universal) when $p$-value is less than 0.05.
	
	\begin{table}[t]
		\caption{Correlation analyses on the graph structure information (\# vertices, \# edges, average degrees, and $dim$) and the optimal $gs$ of the graph}
		\label{tab:correlation}
		\centering
		\tabcolsep 8.5pt
		\vspace{-7pt}
		\begin{tabular}{l|llll}
			\hline
			\multicolumn{1}{c|}{}                             & \multicolumn{4}{c}{Group   sizes}                                                                                                    \\
			\multicolumn{1}{c|}{}                             & \multicolumn{2}{c|}{Bivariate correlation} & \multicolumn{2}{c}{Partial correlation} \\
			\multicolumn{1}{c|}{\multirowcell{-3}{Structure\\information}} & $p$-value                & \multicolumn{1}{l|}{$r$}                     & $p$-value                        & $r$                             \\ \hline
			\# vertices                                        & 0.50218                & \multicolumn{1}{l|}{0.07612}               & 2.28E-04                       & 0.40826                       \\ \hline
			\# edges                                           & 5.66E-04               & \multicolumn{1}{l|}{0.37703}               & 0.00299                        & -0.33405                      \\ \hline
			average   degree                                  & 3.03E-25               & \multicolumn{1}{l|}{0.86648}               & 6.98E-23                       & 0.85306                       \\ \hline
			$dim$                                        & 0.01746                & \multicolumn{1}{l|}{0.26515}               & 0.00369                        & 0.32713                       \\ \hline
		\end{tabular}
		\vspace{-8pt}
		\begin{flushleft}
			{\footnotesize
				$r$ is the correlation coefficient, and $p$-value represents the probability that the correlation in the data occurred by chance.}
		\end{flushleft}
		\vspace{-10pt}
	\end{table}
	
	In Table \ref{tab:correlation}, from the results of the bivariate correlation analysis, we can find that the average degree and the number of edges of the graph are the two metrics that have the strongest (correlation coefficients of 0.87 and 0.38 respectively) correlation with the optimal $gs$. Further, the partial correlation analysis shows that the number of vertices of a graph has a moderate (correlation coefficient of 0.41) correlation with the optimal $gs$ after controlling for the effects of other variables. However, as for the $dim$, the bivariate and partial correlation analyses show that $dim$ has only a \textit{weak} (correlation coefficients of 0.27 and 0.33 respectively) correlation with the optimal $gs$. 
	
	Based on the results of the correlation analyses, we decide to regress the optimal $gs$ for a graph by its \# vertices, \# edges, and the average degree. The regression equation is as follow:
	\begin{equation}
	\begin{aligned}
	gs = &\beta_0 + \beta_1\times \#\ vertices + \beta_2\times \#\ edges \\&+ \beta_3\times average\ degree 
	\end{aligned}
	\end{equation}, where $\beta_0$, $\beta_1$, $\beta_2$, $\beta_3$ are the regression coefficients.
	We then use least squares regression~\cite{pohlmann:least_squares} to calculate the linear regression coefficients and obtain $\beta_0=0.65538$, $\beta_1=1.67431E-5$, $\beta_2=-2.24342E-6$, and $\beta_3=0.63641$.
	
	After obtaining the regression coefficients, for a graph $G$, we just bring in the structure information (i.e., the number of vertices, the number of edges, and the average degree) of $G$ into the regression equation to compute its optimal $gs$. The calculated regression value is rounded to its nearest positive integer as the choice of $gs$ for $G$. 	
	
	\section{Evaluations}\label{sec:eval}
	
	In this section, we first give the experimental setups, and then evaluate our approach from the various angles including space cost, effectiveness on performance improving.
	
	\subsection{Experiment settings} \label{subsec:setup}
	
	\begin{table}[]
		\caption{Graphs used in evaluations (graphs marked by $^*$ are taken from \cite{pygdata}, graphs marked by $^\dagger$ are taken from \cite{snap} and \cite{ryan:NetworkRepository}, Average degree = \# edges$/$\# vertices)}
		\label{tab:dataset}
		\centering
		\tabcolsep 11.5pt
		\vspace{-7pt}
		\begin{tabular}{l|c c c}
			\hline
			Graphs 	&\# vertices 	&\# edges 	&Average degree \\ \hline
			Cora$^*$			& 2,708  			& 5,278 			&1.95 		 \\ 
			Citeseer$^*$		& 3,327  			& 4,552	 			&1.37		 \\ 
			Pubmed$^*$			& 19,717      		& 44,324   			&2.25		 \\ 
			Twitter$^\dagger$	& 81,306			& 1,342,296			&16.51		 \\
			Blog$^\dagger$		& 88,784			& 2,093,195			&23.58		 \\
			Amazon$^\dagger$	& 410,236			& 2,439,437			&5.95 		 \\
			Google$^\dagger$ 	& 875,713			& 4,322,051			&4.94		 \\
			Youtube$^\dagger$	& 1,134,890			& 2,987,624			&2.63		 \\			
			\hline
		\end{tabular}
		%	\vspace{-5pt}
	\end{table}
	
	\textbf{Dataset.} Table \ref{tab:dataset} lists the graphs used in our evaluations. All the graphs are converted to undirected graphs before being used in GCN training. The density (computed by $2\cdot$\# edges$/($\# vertices$\cdot($\# vertices$-1))$) indicates the sparsity of an input graph. Typically, in a semi-supervised task, the training set $V_t$ is a small subset of the vertices of the input graph~\cite{kipf:gcn,engelen:small_set,peikari:small_set,wu:small_set}. If not otherwise specified, we randomly (with uniform distribution) select \textit{10\%} vertices as the training set for each tested graph in the following experiments, i.e., training ratio = 10\%. Note that 10\% training ratio is larger than most of the training ratios in real-world graphs used in GCN-relevant papers.
	
	\textbf{Test-bed.} We implement our proposals (both execution path preparing and structure-aware grouping) by revising the code of GNNAdvisor~\cite{wang:GNNAdvisor}. During revision, we use the structure-aware methods on computing the optimal choice of $gs$ discussed in Section \ref{sec:structure_aware_grouping}, but preserve the method on computing the number of software threads in GNNAdvisor. The prototype system is compiled with CUDA 10.1 and the \texttt{-O3} optimization level. We deploy our prototype system on a Linux server with two Intel Xeon E5-2680 v4 CPUs (14 cores for each, 2.40 GHz), 256 GB memory, and an NVIDIA Tesla P100 GPU~\cite{P100} (3584 cores, 16GB global memory, 64 KB shared memory per streaming multiprocessor). Besides, we evaluate the performances of both GNNAdvisor and DGL when using our approach on an NVIDIA V100S GPU~\cite{V100} (5120 cores, 32GB global memory, 96 KB shared memory per streaming multiprocessor).
	
	\textbf{Baseline.} We choose the performance data of GNNAdvisor~\cite{wang:GNNAdvisor} as the baseline. To be fair, during comparisons, we conduct the same number (100) of epochs of GCN training in both original and improved (using our proposals) GNNAdvisor. 
	
	\subsection{Space cost} \label{subsec:cost_coalesce}
	
	\begin{figure}[t] 
		\centering
		\setlength{\abovecaptionskip}{2pt}
		\includegraphics[width=0.99\linewidth]{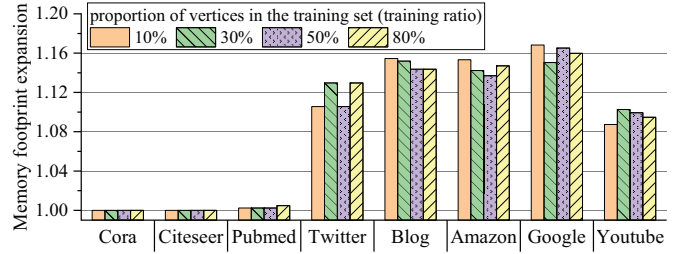}
		\caption{The memory footprints of GCN training by using execution path preparing (normalized to those by using original GNNAdvisor)}
		\label{fig:mem_overhead}
		%	\vspace{-15pt}
	\end{figure}
	
	Our execution path preparing is essentially a \textit{trade-space-for-time} method. Figure \ref{fig:mem_overhead} illustrates our memory footprint expansions by normalizing our memory footprints to those of GNNAdvisor. From Figure \ref{fig:mem_overhead}, we can observe that our execution path preparing incurs only a small fraction (from less than 1\% to about 17\%) of memory expansion compared to GNNAdvisor.
	
	The reason behind such small space costs is that in GCN training problems, compared with feature matrices (i.e., $X^{(l)}$s), weight matrices (i.e., $W^{(l)}$s) and intermediate matrices (e.g., $Y^{(l)}$s and $X^{(l)'}$s in Algorithm \ref{algo:training}), the graph structure data (i.e., the CSRs shown in Figure \ref{fig:group_part}) occupies only a small portion of total memory footprint. Thus, the execution paths (illustrated in Figure \ref{fig:executionpaths}), as subsets of the graph structure data, cause relatively small increments in memory footprint, compared to the memory space consumed by the whole training problem.
	
	Another fact that can be observed from Figure \ref{fig:mem_overhead} is that the space cost incurred by execution path preparing almost remains the same, when the training ratio changes from 10\% to 80\% for the training problems conducted in our chosen graphs. The reason for such a phenomenon is that with the increasing training ratio, the storage space consumed by the execution paths increases accordingly, as more vertices participate in the backward aggregation. However, at the same time, the storage space consumed by GNNAdvisor (our baseline system) also increases, as the intermediate matrices for computing $loss$ and $X^{(L)'}$ enlarge with more training vertices. The memory footprint data in Figure \ref{fig:mem_overhead} suggest that the expansion speed of execution paths due to increasing training ratio does not necessarily exceed that of the intermediate matrices in GNNAdvisor.
	
	\subsection{Performance improvement on backward aggregation} \label{subsec:effect_coalesce}
	
\begin{figure}[t]
		\centering
		\setlength{\abovecaptionskip}{2pt}
		\includegraphics[width=0.99\linewidth]{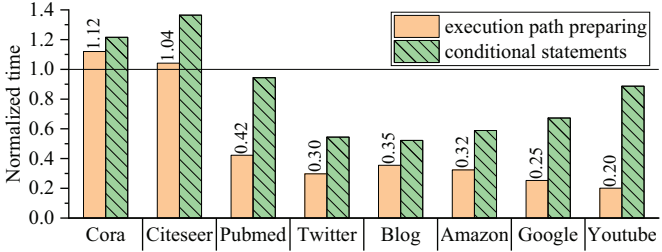}
		\caption{Backward aggregation times by using either execution path preparing or if-else statements (normalized to the times by using original GNNAdvisor)}
		\label{fig:effect_epp}
		%	\vspace{-15pt}
	\end{figure}

	Figure \ref{fig:effect_epp} reports the time paid on backward aggregation using execution path preparing. For comparison, Figure \ref{fig:effect_epp} also depicts the time of backward aggregation using straight forward selective scheduling method, i.e., if-else statements. 
	
	Comparing the performances of backward aggregation using our execution path preparing and those using if-else statements in Figure \ref{fig:effect_epp}, we can observe that the execution path preparing outperforms if-else statements in all training cases on our chosen graphs, while the highest performance improvement (4.41x) appears on \texttt{Youtube}. The reason is that straight-forward selective scheduling method like using if-else statements leads to branch divergence overheads and uncoalesced memory accessing when conducting backward aggregation in the graph data structure, while our execution path preparing technique avoids these overheads and hence leads to better performances.
	
	In Figure \ref{fig:effect_epp}, compared with the original GNNAdvisor, on the left-hand two graphs (i.e., \texttt{Cora} and \texttt{Citeseer}) with only about three thousand vertices, using our execution path preparation alone does not result in better performance. The reason is that when conducting the backward aggregation in Figure \ref{fig:effect_epp}, we use GNNAdvisor's built-in method on computing the group sizes, which leads to severe load imbalance on small graphs and thus degrades performance. However, on the other six graphs (e.g., \texttt{Pubmed}, \texttt{Youtube}), using our execution path preparing alone is able to achieve significant performance improvements (3.4x on average) on backward aggregations.
	
\begin{figure}[t] 
		\centering
		\setlength{\abovecaptionskip}{2pt}
		\includegraphics[width=0.99\linewidth]{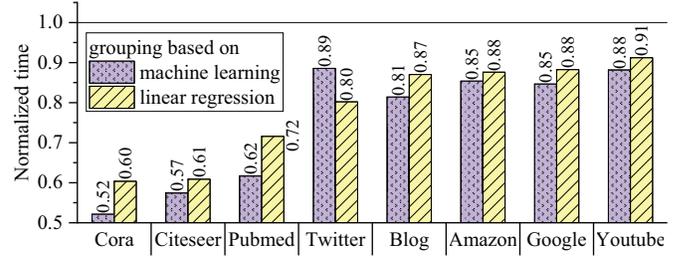}
		\caption{Backward aggregation times by using execution path preparing with structure-aware grouping strategies (normalized to the times by using execution path preparing but with default grouping strategy of GNNAdvisor)}
		\label{fig:tuning_acc}
	\end{figure}

	We further examine the backward aggregation performances by using our structure-aware grouping techniques discussed in Section \ref{sec:structure_aware_grouping}. As we proposed two techniques (machine learning based and linear regression based) on computing the optimal $gs$es for each training layer, Figure \ref{fig:tuning_acc}) presents the backward aggregation performances when using execution path preparing with these two grouping techniques respectively, and compares the performances with those (presented in Figure \ref{fig:effect_epp}) when using execution path preparing with grouping strategies of GNNAdvisor.
	
	From Figure \ref{fig:tuning_acc}, we can observe that both our machine learning based and linear regression based methods on computing $gs$es are effective in improving the performance of backward aggregations, and lead to relatively large performance speedups on relatively small graphs like \texttt{Cora} and \texttt{Citeseer}, where execution path preparing alone does not achieve prominent acceleration as shown in Figure \ref{fig:effect_epp}. The reason is that on computing optimal $gs$es during GCN training, GNNAdvisor considers only the dimension property of training at each layer, and does not lead to optimal performances for small graphs as discussed in Section \ref{sec:structure_aware_grouping}. But our proposed methods achieve better performances as we consider the structure of the input graphs. 
	
	Comparing our two grouping methods, it can be seen that the machine learning based method always leads to shorter backward aggregation times on all tested graphs except \texttt{Twitter}. However, the machine learning based method takes a much longer time (details in Table \ref{tab:net_overall}) to compute the optimal $gs$ for an input graph, while the linear regression based method has negligible computation time of the $gs$ value. Therefore, the machine learning based method is preferable when seeking the shortest aggregation time, or when the input graph is used for many runs of the GCN training and each run involves thousands of epochs~\cite{li:deepergcn,william:graphsage}. And the linear regression based method can lead to a shorter end-to-end time than the machine learning based method when conducting small tests such as one run of 100 epochs.
	
	Combining the performance data shown in Figure \ref{fig:effect_epp} and Figure \ref{fig:tuning_acc}, we can obtain the effectiveness of our proposed execution path preparing on the performances of backward aggregation, when accompanied with the techniques on computing the optimal group size. We can observe that for all our chosen graphs, our proposal is effective in accelerating the backward aggregations with varying degrees. The smallest speedup appears on the \texttt{Cora} graph, where the performance speedup is 1.48x, computed by $(1/1.12)\times(1/0.60)$ (speedup by execution path preparing $\times$ speedup by linear regression based grouping). The biggest speedup happens in the \texttt{Youtube} graph, where the performance speedup is 5.68x, computed by $(1/0.20)\times(1/0.88)$ (speedup by execution path preparing $\times$ speedup by machine learning based grouping).
	
	\begin{figure}[t]
		\centering
		\setlength{\abovecaptionskip}{2pt}
		\includegraphics[width=0.99\linewidth]{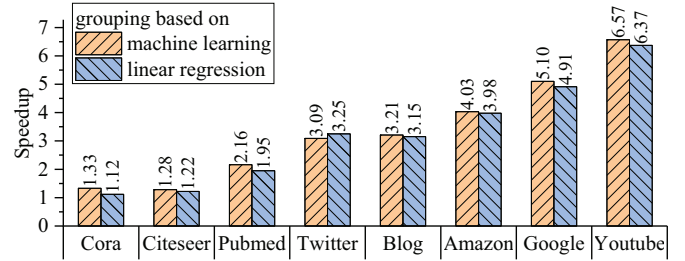}
		\caption{Speedups by using our approach compared with original GNNAdvisor on backward aggregation stage on NVIDIA Tesla V100S GPU}
		\label{fig:speedups_v100s}
	\end{figure}
	
	\textbf{Performance on NVIDIA Tesla V100S:} We use our approach to improve the backward aggregation of both GNNAdvisor and DGL on an NVIDIA Tesla V100S GPU. 
	
	$\bullet$ Improving GNNAdvisor: Figure \ref{fig:speedups_v100s} reports the speedups of the improved GNNAdvisor using our approach over its original version. We can observe that our approach is still effective in improving the performance of GNNAdvisor (speedups range from 1.12x to 6.57x) when the GPU device is upgraded from P100 to V100S. It is noticeable that the speedups are higher (especially for \texttt{Amazon}, \texttt{Google}, and \texttt{Youtube}) on V100S than those on P100. The reason is that, when deciding the optimal $gs$es for input graphs, GNNAdvisor (considers only the platform parameters and) does not take into account the structural information of the graph, while our approach (discussed in Section \ref{sec:structure_aware_grouping}) does better on computing the optimal $gs$es.
	
	$\bullet$ Improving DGL: DGL uses SpMM to conduct backward aggregation of GCN training. From the angle of graph processing, SpMM in DGL during backward aggregation is still an all-active procedure, since the same adjacency matrix of input graph is used for both forward and backward aggregations. Hence, we can also use our execution path preparing approach to improve the backward aggregation performance of DGL. The method is to represent the execution paths as matrices (with the same representation of the adjacency matrix of input graph, but much smaller), to convert a big SpMM problem into multiple smaller SpMM problems during backward aggregation. Table \ref{tab:com_dgl} lists the execution times of backward aggregation of GCN trainings in our chosen graphs with DGL, before and after improvements.
	
	From Table \ref{tab:com_dgl}, we can observe that our approach is also effective in improving the performance of backward aggregation in DGL. Nevertheless, the speedups in DGL are generally smaller (1.02x$\sim$2.78x) than those in GNNAdvisor. The reason is that DGL uses SpMM and does not use the grouping techniques, the technique on computing optimal group sizes discussed in Section \ref{sec:structure_aware_grouping} cannot be used in such improvements. Comparing the absolute execution times of GNNAdvisor and DGL, both before and after improvements, we can observe that although in some (\texttt{Amazon}, \texttt{Google}, and \texttt{Youtube}) graphs, the backward aggregation performances of original DGL are better than those of original GNNAdvisor, they are much worse than those of GNNAdvisor after both of them being improved. The reason is that although the latest version (0.9.0) of DGL uses customized SpMM libraries to conduct training, and achieves even better performances than GNNAdvisor in some graphs, it leaves little room except for using smaller matrices (representing the sub-graphs) in backward aggregation for our approach to further improve its performance. 
	
	\begin{table}[]
		\caption{Backward aggregation times (in seconds) of DGL (version 0.9.0) and GNNAdvisor before and after being improved.}
		\label{tab:com_dgl}
		\centering
		\tabcolsep 4pt
		\vspace{-7pt}
		\begin{tabular}{c|ccc||cc}
			\hline
			\multirow{2}{*}{Graphs} & \multicolumn{2}{c}{DGL}  & \multirow{2}{*}{Speedup} & \multicolumn{2}{c}{GNNAdvisor} \\
			& Original  & Improved   &   & Original  & Improved\\ \hline
			Cora     & 0.1506 & 0.1476    & 1.02    & 0.0038 & 0.0034   \\ 
			Citeseer & 0.1534 & 0.1504    & 1.02    & 0.0039 & 0.0032   \\ 
			Pubmed   & 0.1546 & 0.1501    & 1.03    & 0.0084 & 0.0043   \\ 
			Twitter  & 0.1850 & 0.1652    & 1.12    & 0.1098 & 0.0338   \\ 
			Catalog  & 0.2474 & 0.2028    & 1.22    & 0.1572 & 0.0499   \\ 
			Amazon   & 0.2860 & 0.1497    & 1.91    & 0.3274 & 0.0823   \\ 
			Google   & 0.3648 & 0.1681    & 2.17    & 0.6101 & 0.1243   \\ 
			Youtube  & 0.6239 & 0.2244    & 2.78    & 0.7659 & 0.1203   \\ \hline
		\end{tabular}
	\end{table}
	
	\subsection{Performance improvement on overall GCN training}
	
	In this subsection, we examine the effectiveness of our proposed method (execution path preparing combined with structure-aware grouping techniques) on improving the performances of overall GCN training. According to the overheads of our proposed techniques on execution path preparing and computing the optimal group sizes, we consider two scenarios that use our proposed method: during pre-processing and on-the-fly.
	
	\textbf{During pre-processing:} In this scenario, the pre-processing first extract the execution paths of the backward aggregation before the GCN training, then computes the optimal group sizes by using the SAGPG neural network as discussed in Section \ref{sec:structure_aware_grouping}, and finally reconstructs the execution paths by the computed group sizes. After pre-processing, the GCN training is conducted while the reconstructed execution paths are used during its backward aggregation.
	
	Table \ref{tab:net_overall} lists the pre-processing times and the training times (in bold) separately. From the table, we can observe that when comparing only the training times with those of GNNAdvisor, our approach is effective in improving the performances of overall training. The speedups range from 1.05x to 1.37x, where small speedups appear on three small graphs (i.e., \texttt{Cora}, \texttt{Citeseer}, and \texttt{Pubmed}), for the proportions of backward aggregation in overall training on these small graphs are relatively small (about 20\%). 
	
	\begin{table}[t]
		\caption{The overall training times (in seconds) of GNNAdvisor and the pre-processing version of our approach (``Accuracy difference'' is computed by subtracting the Accuracy from GNNAdvisor with those from our approach)}
		\vspace{-7pt}
		\label{tab:net_overall}
		\centering 
		\tabcolsep 3.5pt
		\begin{tabular}{c|cc|c|c|c}
			\hline
			\multirow{2}{*}{Graph}	& \multicolumn{2}{c}{Our approach} & GNNAdvisor & \multirow{2}{*}{Speedup}  & Accuracy  \\
			& Pre-proc.	& Training 	& Training	& & difference\\ \hline
			Cora	& 0.1591			& \textbf{0.1999}	& \textbf{0.2234}	& 1.12 	& 0.000\\
			Citeseer& 0.1640			& \textbf{0.2253}	& \textbf{0.2374}	& 1.05	& 0.000\\
			Pubmed	& 0.2236			& \textbf{0.2396}	& \textbf{0.2640}	& 1.10	& 0.000\\
			Twitter & 0.7524			& \textbf{0.6333}	& \textbf{0.8652}	& 1.37	& 0.000\\
			Blog 	& 0.8578			& \textbf{0.8266}	& \textbf{1.1051}	& 1.34	& 0.000\\
			Amazon  & 7.1804			& \textbf{1.8244}	& \textbf{2.3003}	& 1.26	& 0.000\\
			Google  & 22.1762			& \textbf{3.5601}	& \textbf{4.5226}	& 1.27	& 0.000\\
			Youtube & 29.6670			& \textbf{4.9308}   & \textbf{6.0142}	& 1.22  & 0.000\\
			\hline              
		\end{tabular}
		%\vspace{-10pt}
	\end{table}

	Moreover, from Table \ref{tab:net_overall}, we can also observe that the times paid on pre-processing are higher than those paid on training, and larger input graphs lead to higher pre-processing time to training time ratios. The reason is that the majority (99\%) of pre-processing times are spent on the inference of SAGPG. The neural network model of SAGPG is more complex and deeper (i.e., 9 according to~\cite{junhyun:SAG}) than that of GCN, which explains the higher pre-processing times and their faster increasing when processing larger input graphs, than those of GCN training. Based on this observation, we believe that the pre-processing scenario fits the case where the training problem is fixed ($G$, $V_t$, and $L$ do not change), but the training is performed multiple times, and each time contains thousands of epochs.
	
	In addition, we compare the Accuracy (when training with the same random seeds) of our approach and those of GNNAdvisor by computing their differences in the last column of Table \ref{tab:net_overall}. We can observe from the column that our execution path preparing approach does not cause loss of Accuracy as proved in Section \ref{sec:partially_active}. As further validation, our accuracies on the three citation graphs \texttt{Cora}, \texttt{Citeseer}, and \texttt{Pubmed} are 81.2\%, 71.2\%, and 79.5\%, respectively, which are consistent with the results in \cite{kipf:gcn}.
	
	\textbf{On-the-fly:} In this scenario, the execution path preparing is conducted during the first epoch of training (before the backward aggregation stage). Also before backward aggregation, the linear regression based method is adopted to compute the optimal group sizes for each execution path at each training layer.
	
	\begin{figure}[t]
		\centering
		\setlength{\abovecaptionskip}{2pt}
		\includegraphics[width=0.99\linewidth]{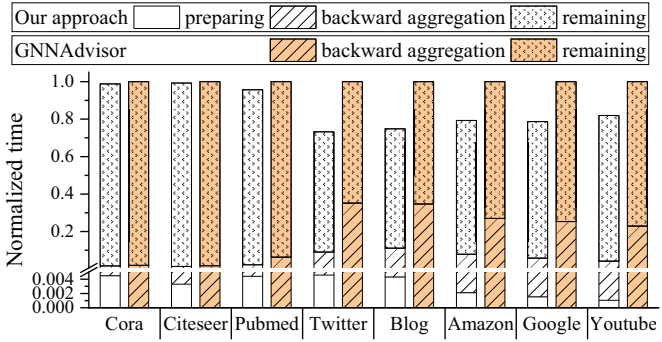}
		\caption{Overall training times by using our approach on-the-fly (normalized to the training times of original GNNAdvisor, and the remaining part includes the times of forward propagation, combinations, and other misc. operations)}
		\label{fig:map_overall}
	\end{figure}
	
	Figure \ref{fig:map_overall} breakdowns the training, and illustrates the execution times of different stages of the training by normalizing them to those GNNAdvisor. From Figure \ref{fig:map_overall}, we can observe that by only improving the performances of backward aggregation, our execution path preparing approach is effective in improving the overall performances of GCN training when used on-the-fly, and the speedups range from 1.03x to 1.35x.
	
	\subsection{Training with more than two layers}
	
\begin{figure}[t]  
		\centering
		\setlength{\abovecaptionskip}{3pt}
		\captionsetup[subfloat]{captionskip=0pt}
		\subfloat[][Memory expansions]{
			\label{fig:ml_mem}
			\includegraphics[width=0.48\linewidth]{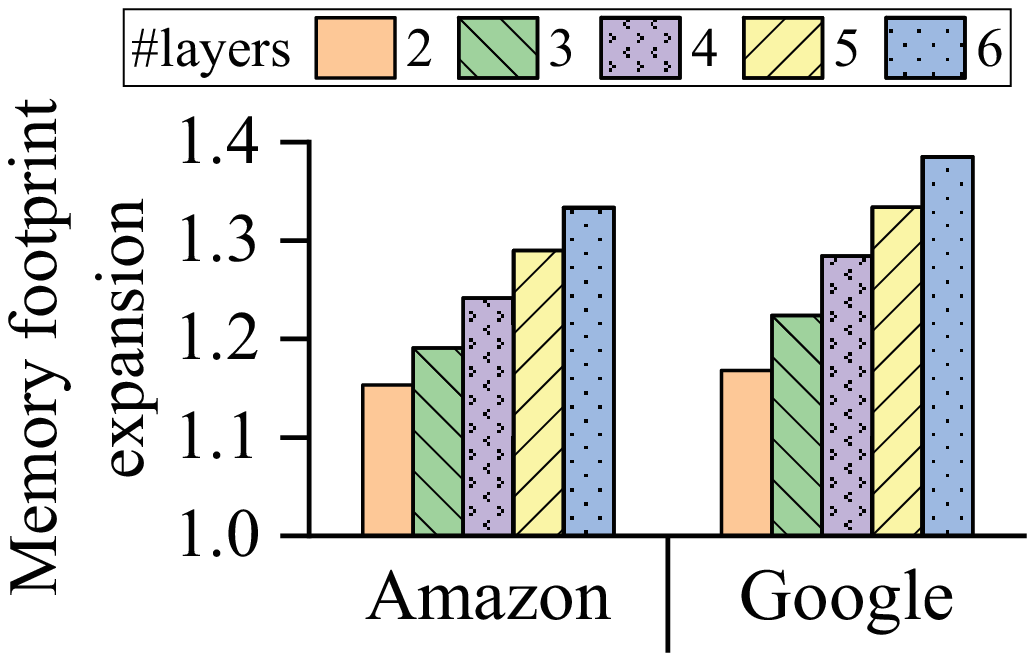}
		}
		\hfill
		\subfloat[][Speedups]{
			\label{fig:ml_speedup}
			\includegraphics[width=0.419\linewidth]{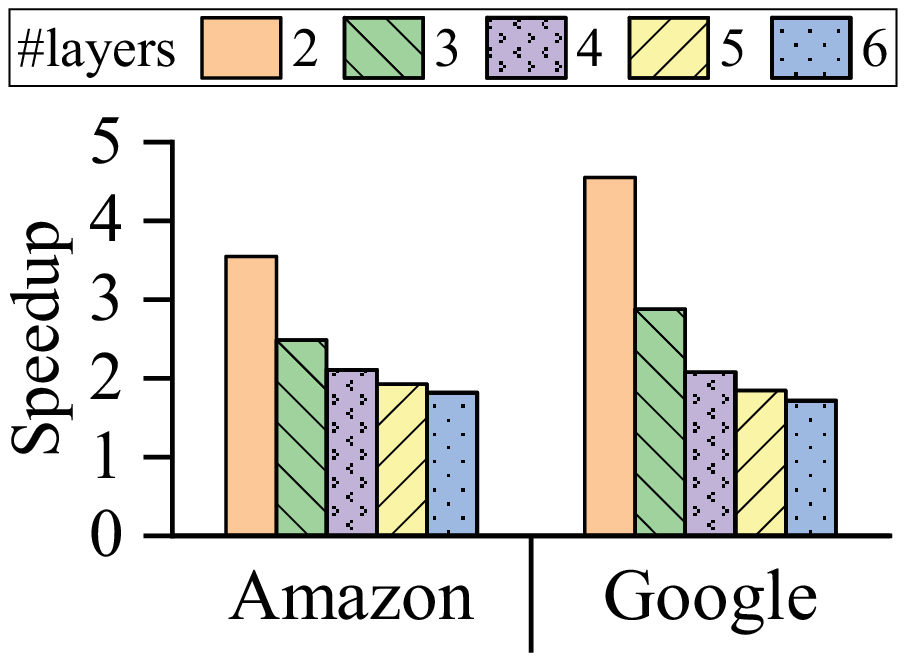}
		}
		
		\caption{Performances of GCN training with 2$\sim$6 layers by using execution path preparing and regression-based grouping (normalized to performances of original GNNAdvisor)}
		\label{fig:multi_layer}
	\end{figure}

	We evaluate the space costs and performance improvements on backward aggregation when $L>2$ by using execution path preparing, where linear regression based method is employed in computing the $gs$es. Figure \ref{fig:multi_layer} reports the experiment results when conducting GCN training on two typical real-world graphs (i.e., \texttt{Amazon} and \texttt{Google}). 
	
	From Figure \ref{fig:multi_layer}, we can observe that with increasing training layers, the space costs increase gradually, and each increasing training layer leads to about a 5\% increment in the memory footprint for both of our chosen graphs. This is because our execution path preparing approach needs to extract data from the original graph, and build the sub-graph (i.e., execution path) for the increased training layer. Moreover, it is likely for large training layers to obtain the majority of data (vertices and edges) of the original graph, according to the small-world theory~\cite{clauset:powerlaw} in real-world graphs. We can also observe that the performance speedups on backward aggregation gradually decrease with more training layers. The reason is that at high training layers, the execution path is close to the original graph by including the majority of its vertices and edges, the performance benefits from selective scheduling thus diminishes.
	
	\section{Conclusion and Future Work}\label{sec:conclusion}
	
	In this paper, we show that the backward aggregation stage of GCN training uses only partial data of its input graph, and thus can be treated as a partially-active graph processing procedure. By leveraging this finding, we propose an execution path preparing approach to improve the performance of backward aggregation stage. Experiment results show that with a small space cost, our proposal is effective in improving the performance of backward aggregation, and the overall performance of GCN training. 
	
	The future work of this paper includes: 1) practically applying our proposal to deep GCNs~\cite{rong:deep} (GCNs with more than two layers); 2) extending our proposal to other partially active situations such as the aggregations after graph pooling~\cite{mesquita:pooling} or sampling~\cite{william:graphsage}, both of which select a subset of vertices in the input graph to participate in the following computation in forward propagation.

	\bibliographystyle{IEEEtran}
	\bibliography{IEEEabrv,reference}
	
\end{document}